\documentclass[letterpaper, 10 pt, conference]{ieeeconf}  

\IEEEoverridecommandlockouts                              

\usepackage{graphicx} 
\usepackage{caption} 

\usepackage{algorithm}
\usepackage{algorithmic}
\usepackage{booktabs} 
\usepackage{multirow}
\usepackage{bm}
\usepackage[table]{xcolor}
\usepackage{amsfonts}

\newcommand{\R}{\mathbb{R}}

\newcommand{\bg}{\mathbf{g}}
\newcommand{\tok}{\mathrm{t}}

\title{\LARGE \bf
Sparse View Distractor-Free Gaussian Splatting
}

\author{Yi Gu$^{1,*}$, Zhaorui Wang$^{1,*}$,  Jiahang Cao$^{1}$, Jiaxu Wang$^{1}$, Mingle Zhao$^{2}$, Dongjun Ye$^{1}$, Renjing Xu$^{1 \dagger}$ 
\thanks{* Authors contributed equally to this work.}
\thanks{$^{1}$ The Hong Kong University of Science and Technology (Guangzhou)}%
\thanks{$^{2}$ University of Macau}%
\thanks{$\dagger$ Corresponding author
        {\tt\small renjingxu@hkust-gz.edu.cn}}%
}

\begin{document}

\maketitle
\thispagestyle{empty}
\pagestyle{empty}

\begin{abstract}
3D Gaussian Splatting (3DGS) enables efficient training and fast novel view synthesis in static environments. To address challenges posed by transient objects, distractor-free 3DGS methods have emerged and shown promising results when dense image captures are available. However, their performance degrades significantly under sparse input conditions. This limitation primarily stems from the reliance on the color residual heuristics to guide the training, which becomes unreliable with limited observations. In this work, we propose a framework to enhance distractor-free 3DGS under sparse-view conditions by incorporating rich prior information. Specifically, we first adopt the geometry foundation model VGGT to estimate camera parameters and generate a dense set of initial 3D points. Then, we harness the attention maps from VGGT for efficient and accurate semantic entity matching. Additionally, we utilize Vision-Language Models (VLMs) to further identify and preserve the large static regions in the scene. We also demonstrate how these priors can be seamlessly integrated into existing distractor-free 3DGS methods. Extensive experiments confirm the effectiveness and robustness of our approach in mitigating transient distractors for sparse-view 3DGS training.
\end{abstract}

\section{Introduction}
\label{introduction}

Photorealistic 3D reconstruction from multi-view images are fundamental problem in computer vision and computer graphics, with a broad range of robotics-related applications, e.g., virtual reality~\cite{zhu2025vr}, motion planning~\cite{xu2024hgs,liu2025physics,jin2024gs}, navigation~\cite{yamazaki2024open,jun2024renderable}, etc. Neural Radiance Fields (NeRF)~\cite{mildenhall2021nerf} have achieved significant breakthroughs by leveraging implicit representations and volume rendering~\cite{max2002optical} to improve view synthesis quality. 3D Gaussian Splatting (3DGS)~\cite{kerbl20233d} introduces an explicit point-based representation that supports fast training and enables high-quality, real-time rendering. Despite these advances, 3DGS is primarily designed for static scenes and typically relies on densely captured images to reconstruct detailed 3D geometry.

Existing studies have explored sparse-view reconstruction by incorporating co-visibility and geometry constraints~\cite{zhang2024cor, li2024dngaussian} or leveraging data-driven feedforward approaches~\cite{chen2024mvsplat, ye2025nop, zhang2025FLARE}, primarily targeting static scenes. Another line of research aims to extend NeRF or 3DGS to dynamic environments where transient objects are present. Most distractor-free methods rely on explicit or implicit color residual heuristics~\cite{sabour2023robustnerf, chen2024nerf} to guide the training process. However, such heuristics tend to be less effective under sparse-view conditions, raising an important question: Can transient information be identified before the optimization?

In this work, we investigate how to extract rich mask prior information to enhance sparse-view 3D reconstruction. Given the superiority of 3DGS, we adopt it as our primary representation and demonstrate how our method facilitates distractor-free 3DGS training. To train a 3DGS model, the conventional data preparation pipeline typically relies on Structure from Motion (SfM) tools such as COLMAP~\cite{schonberger2016structure} to estimate accurate camera parameters and generate sparse points. However, this approach often fails in the sparse-view setting or scenes with significant transient contents. Even when image registration succeeds, the resulting point cloud is usually too sparse to support high-quality reconstruction.

To improve the robustness of data preparation, we leverage the geometry foundation model VGGT~\cite{wang2025vggt} to estimate camera parameters. VGGT not only runs significantly faster but also produces a dense initial point cloud. However, this initial point cloud often includes many transient objects, complicating downstream processing. Interestingly, we observe that the attention maps in VGGT are highly informative and capable of consistently identifying the same objects across views without manual supervision. We exploit this property to automatically match corresponding objects and use the geometric outputs from VGGT to further validate the matching pairs. As a result, we can assign each image a prior static mask, where the unmasked regions correspond to potentially transient contents.

To improve the accuracy of prior masks, we investigate the potential of powerful Vision-Language Models (VLMs) to assist in our task. We encode our objectives into text prompts and directly input them alongside the images into VLMs. However, we observe that modern VLMs often misinterpret the task and may hallucinate irrelevant regions. To mitigate these issues, we simplify the problem by restricting VLM queries to large unmasked areas only. Experimental results show that our method generates highly accurate prior masks, significantly outperforming baseline approaches.

To demonstrate how these priors can be seamlessly integrated into existing distractor-free 3DGS framework, we incorporate them into RobustGS~\cite{ungermann2024robust}. We introduce a simple yet effective warm-up strategy guided by mask priors, where the initial training masks in GS are replaced with our prior masks. Experimental results show that this approach yields substantial improvements, including a 1–4 dB gain in PSNR and significantly enhanced handling of distractors.

Our key contributions are as follows:
\begin{itemize}
    \item We propose an efficient and robust method for generating static masks by leveraging attention maps from VGGT to match corresponding mask pairs.
    \item We explore the use of Vision-Language Models for the distractor-free reconstruction and introduce a reliable prompting strategy to generate high-quality prior masks.
    \item We demonstrate how these prior masks can be seamlessly integrated into existing distractor-free 3DGS frameworks and achieve state-of-the-art results.
\end{itemize}

\section{Related Work}
\noindent
\textbf{Distractors handling in NeRF and 3DGS.} The original NeRF \cite{mildenhall2021nerf} relies on strong assumptions about the capture setup, requiring the scene to remain perfectly static and the lighting conditions to stay consistent throughout the capture. Two pioneer works, NeRF-W~\cite{martin2021nerf} and RobustNeRF~\cite{sabour2023robustnerf}, extend NeRF to unstructured ``in-the-wild'' captured images by using photometric error as guidance. RobustNeRF approached the problem from a robust estimator perspective, with binary weights determined by thresholded rendering error, and a blur kernel to reflect the assumption that pixels belonging to distractors are spatially correlated. However, both RobustNeRF and NeRF-W variants rely solely on RGB residual errors, which often leads to misclassification of transients that share similar colors with the background. NeRF-HuGS~\cite{chen2024nerf} combines heuristics from COLMAP’s sparse point cloud and off-the-shelf semantic segmentation to remove distractors, but both heuristics are shown to fail under heavy transient occlusions. NeRF On-the-go~\cite{Ren2024NeRF} uses semantic features from DINOv2~\cite{oquab2024dinov2} to predict outlier masks via a lightweight MLP. However, it also relies on direct supervision from structural rendering errors, leading to potential over- or under-masking of outliers.

Following the evolution of Distractor-free NeRF methods, multiple works~\cite{sabour2025spotlesssplats, ungermann2024robust} also use color residual heuristics as the transients guidance in the Gaussian Splatting. Benefiting from the explicit properties, some works~\cite{lin2024hybridgs, wang2024desplat} propose to use view-specific Gaussian points for modeling per-view distractors, which utilize color residual heuristics implicitly. Extending the robustness to unconstrained photo collections, RobustSplat~\cite{fu2025robustsplat} and RobustSplat++~\cite{fu2025robustsplat++} introduce uncertainty modeling to mask out transient occluders alongside appearance variations. Similarly, DeGauss~\cite{wang2025degauss} proposes a decomposed framework that decouples dynamic entities from static scenes by leveraging semantic feature fields. Other works~\cite{kulhanek2024wildgaussians, park2025forestsplats} also incorporate DINOv2~\cite{oquab2024dinov2} or Stable Diffusion~\cite{rombach2022high} feature priors~\cite{sabour2025spotlesssplats} to handle occlusions. In this work, we focus on unordered sparse images with minimal appearance changes. Thus, we exclude methods that primarily address appearance changes~\cite{zhang2024gaussian, xu2024wild} or rely on temporal information~\cite{goli2024romo, markin2024t, xu2024das3r}.

\noindent
\textbf{3DGS with geometry foundation models.} While neural 3D reconstruction has made significant progress, it often relies on densely captured multi-view inputs with carefully initialized camera poses, typically obtained via Structure-from-Motion (SfM) tools such as COLMAP~\cite{schonberger2016structure}. This dependence limits its applicability in real-world scenarios, where sparse-view inputs and limited feature matches can lead to unreliable pose estimation and cumulative reconstruction errors. To address these challenges, InstantSplat~\cite{fan2024instantsplat} introduces a novel and extremely fast neural reconstruction pipeline built upon the Geometry Foundation models DUSt3R~\cite{wang2024dust3r}/MASt3R~\cite{leroy2024grounding}, capable of recovering accurate 3D representations from as few as 2–3 input images. However, InstantSplat remains focused on static scenes. While concurrent SparseGS-W~\cite{li2025sparsegs} leverages DUSt3R and diffusion models~\cite{rombach2022high} for initialization and enhancement, it relies on manual, user-defined text prompts (e.g., "tourist") to generate mask priors.
In contrast, our method is a more general and fully automated process that eliminates the need for pre-defined scene-specific prompts. A closely related effort is Easi3R~\cite{chen2025easi3r}, which also explores the role of attention mechanisms in DUSt3R. Unlike Easi3R, our method is independently developed and specifically targets unordered, sparsely captured images without assuming any available motion or temporal consistency.


\section{Preliminaries}
\textbf{3DGS}~\cite{kerbl20233d} explicitly represents 3D scenes with a set of 3D Gaussians $\{\mathcal G_i\}$. Each Gaussian is defined by a Gaussian function:
$$
\mathcal G_i(\bm{x}|\bm{\mu}_i, \bm{\Sigma_i}) = e^{-\frac{1}{2}(\bm{x} - \bm{\mu}_i)^\top\bm{\Sigma}_i^{-1}(\bm{x}-\bm{\mu}_i)},
$$
where $\bm{\mu}_i \in \mathbb R^3$ and $\bm{\Sigma}_i \in \mathbb R^{3\times3}$ are the center of a point $\bm{p}_i\in \mathcal P$ and corresponding 3D covariance matrix, respectively. The covariance matrix $\bm{\Sigma}_i$ can be decomposed into a scaling matrix $\bm{S}_i\in \mathbb R^{3\times3}$ and a rotation matrix $\bm{R}_i\in \mathbb R^{3\times3}$:
$$
\bm{\Sigma}_i = \bm{R}_i\bm{S}_i\bm{S}_i^\top \bm{R}_i^\top.
$$

3DGS allows fast $\alpha$-blending for rendering. Given a transformation matrix $W$ and an intrinsic matrix $\bm{K}$, $\bm{\mu_i}$ and $\bm{\Sigma}_i$ can be transformed to camera coordinate corresponding to $\bm{W}$ and then projected to 2D coordinates:
$$
\bm{\mu}_i^{'}=\bm{KW}[\bm{\mu}_i,1]^\top, \quad  \bm{\Sigma}_i^{'}=\bm{JW\Sigma}_i\bm{W}^\top \bm{J}^\top,
$$
where $J$ denotes the Jacobian matrix of the projective transformation. Rendering color $\bm{C}\in\mathbb R^3$ of a pixel $\bm u$ can be obtained in a manner of $\alpha$-blending:
$$
\bm{C} = \sum_{i\in N_{\mathcal G}} T_i \alpha_i \bm{c}_i,\quad T_i=\prod_{j=1}^{i-1}(1 - \alpha_i),
$$
where $\alpha_i$ is calculated by evaluating $\mathcal G_i(\bm{u}|\bm{\mu}_i^{'}, \bm{\Sigma}_i^{'})$ multiplied with a learnable opacity corresponding to $\mathcal G_i$, and the view-dependent color $\bm{c}_i\in\mathbb R^3$ is represented by spherical harmonics (SH) from the Gaussian $\mathcal G_i$. $T_i$ is the cumulative opacity. $N_\mathcal G$ is the number of Gaussians that the ray passes through.

\textbf{VGGT}~\cite{wang2025vggt} is a feed-forward neural network that directly infers camera parameters, point maps, depth maps, and point tracks, from multiple views. Specifically, given $N$ RGB images $(I_i)_{i=1}^N$, $I_i \in \R^{3\times H\times W}$. VGGT's transformer is a function that maps this sequence to a corresponding set of 3D annotations, one per frame:
\begin{equation}
f\left((I_i)_{i=1}^N\right) = \left(\bg_i, D_i, P_i, T_i\right)_{i=1}^N.
\end{equation}
$\bg_i \in \mathbb{R}^9$ is the camera parameters of image $I_i$. The world reference frame is defined in the coordinate system of the first camera $\bg_1$. The depth map $D_i \in \R^{H\times W}$, and point map $P_i \in \R^{3\times H\times W}$ are redundancy designs to ease the training. Unless mentioned otherwise, we only use the depth map $D_i$ to estimate dense geometry, which usually leads to more accurate 3D points than the point map branch.


The input image $I_i$ of VGGT is initially patchified into a set of $K$ tokens where $\tok^I_i \in \R^{K\times C}$. The combined set of image tokens from all frames, i.e., $\tok^I = \cup_{i=1}^N \{\tok^I_i\}$, is subsequently processed through alternating frame-wise and global self-attention layers~\cite{vaswani2017attention}. The \textit{frame-wise self-attention} learns the monocular geometry by attending to the tokens $\tok^I_k$ within each frame separately. The \textit{global self-attention} attends to the tokens $\tok^I$ across all frames jointly, which aggregates information from multiple views without any inductive bias.

\section{Method}
\subsection{Method Overview}
Given $N$ color images, we first employ VGGT to obtain the initial point cloud and camera parameters. Then, we use a class-agnostic mask predictor to process each image and derive the 2D masks $\{\bm{m}_{i, k} | k = 1, ..., n_i\}$, where $n_i$ denotes the number of masks in the image $I_i$. As presented in previous works~\cite{ungermann2024robust}, SAM~\cite{kirillov2023segment} tends to over-segment the image. In this work, we use CropFormer~\cite{qi2023high} as our mask predictor, since the entity-level segmentation is well suited for the distractor-free task~\cite{otonari2024entity}. Considering the color residual heuristics are not reliable in the sparse-view setting, we decide to incorporate rich mask prior information before the GS training. Specifically, we utilize the attention map from VGGT to achieve fast entity matching (Sec.~\ref{attention_sec}). After that, we resort to the VLM model to identify and preserve the large static regions in the scene (Sec.~\ref{VLM}). We demonstrate how to apply our mask priors to existing distractor-free GS frameworks in Sec.~\ref{Integrate}.

\begin{figure}[!htbp]
\centering
\includegraphics[width=0.98\linewidth]{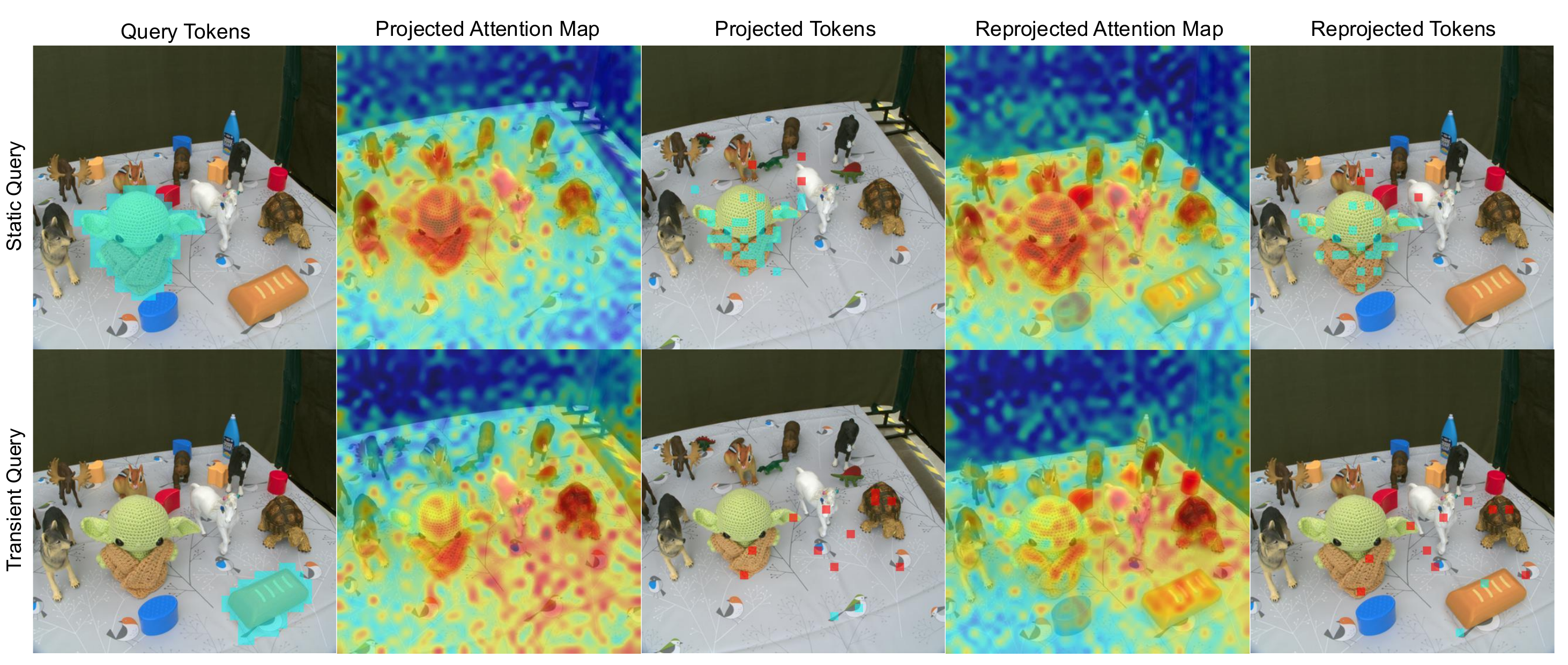}
\definecolor{custom_cyan}{rgb}{0.42,0.91,0.91}
\vspace{-0.3cm}
\caption{
Illustration of VGGT attention-guided semantic entity matching. Query tokens are highlighted in \textbf{\textcolor{custom_cyan}{cyan}}. Initially, we project these query tokens onto the reference image to obtain the projected tokens. The reprojected tokens are computed in a similar manner. A projected token is considered valid only if its reprojected counterpart lies within the region of the query tokens; otherwise, it is classified as an invalid token (colored in \textbf{\textcolor{red}{red}}). To explicitly illustrate the cross-view correspondence matching process, we visualize the global feature maps. Compared to static objects, transient objects typically exhibit lower recall, which serves as a primary criterion for identifying distractors.
}
\label{fig:vggt_atth} 
\vspace{-0.4cm}
\end{figure}
\subsection{VGGT-attention Guided Entity Matching}
\label{attention_sec}
Our primary objective is to estimate the static mask priors prior to 3DGS training. Building on the assumption from NeRF-HuGS~\cite{chen2024nerf}, the observation frequency of static objects should be higher than that of transient objects. However, relying on SfM-based heuristics, as proposed in NeRF-HuGS, may be less reliable in sparse-view settings due to the limited number of matching pairs and the presence of numerous outliers. In the context of VGGT-based initialization, geometry reprojection methods are constrained by the accuracy of VGGT. Alternatively, we can leverage the VGGT tracking head to achieve this goal. However, tracking too many points proves inefficient, and determining the confidence and visibility thresholds for VGGT tracking results remains challenging. Semantic-level matching, as proposed in SpotLessSplat~\cite{sabour2025spotlesssplats}, may be less effective in the presence of repetitive objects. Based on these observations, we aim to address this issue at both the geometric and semantic levels.

VGGT learns the high-level feature matching implicitly in the \textit{global self-attention} layers. Given a pair of images, the query image $I_i$ and the reference image $I_j$, where $j \neq i$. We extract the query tokens corresponding to the mask $\bm{m}_{i, k}$ according to whether the patch is occupied by the mask. By setting the number of query tokens in each layer as $S$, the query tokens can be represented as $\tok^I_{i, k, l} \in \mathbb{R}^{S\times C_f}$, where $C_f$ is the feature dimension and $l$ is the layer index. Then, we use $\tok^I_{i, k}$ to attend $\tok^I_j$:
\begin{equation}
    \mathcal{A} = \tok^I_{i, k} {\tok^I_j}^T / \sqrt{C_f}.
\end{equation}
Here we reshape $\mathcal{A}$ as $\mathcal{A} \in \mathbb{R}^{S \times L \times h \times w}$, where $L$ is the number of global attention layers and $h, w$ are the height and width of feature maps. By averaging the layer dimension, we will get an aggregated attention map $\bar{\mathcal{A}} \in \mathbb{R}^{S \times h \times w}$. Then, we obtain the best matching tokens by selecting the index in $I_j$  with the highest attention value:
\begin{equation}
    Index_{j, s} = \arg \max_{s\in\{1, ..., S\}}(\bar{\mathcal{A}}).
\end{equation}
This operation can be viewed as a projection process, which considers the relevance at both feature and geometric levels. By index on $Index_{j, s}$, we can obtain the projected tokens $\tok^{proj_{i,k}}_{j}$.

To filter the noised unrelated tokens in $\tok^{proj_{i,k}}_{j}$, we re-project the $\tok^{proj_{i,k}}_{j}$ to attend the image $I_i$.
The re-projected tokens are represented as $\tok^{rep_{j}}_{i}$. 
\begin{figure}[t] 
\centering
\includegraphics[width=0.98\linewidth]{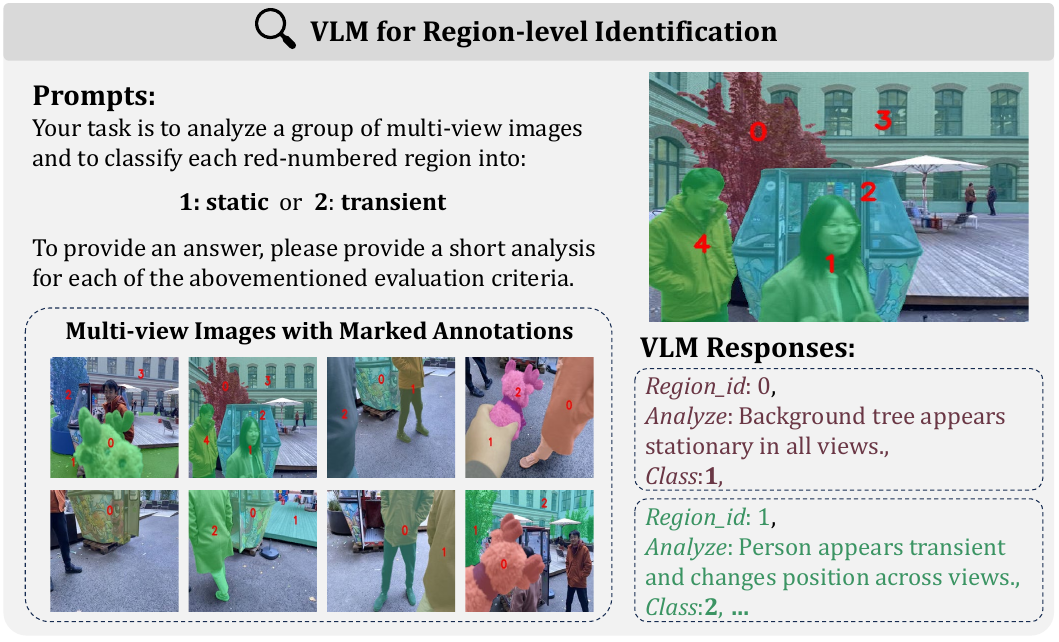}
\vspace{-0.3cm}
\caption{
VLM process illustraion. To simplify annotations, we exclude masks containing fewer than 20000 pixels. For the remaining transient candidate masks, we automatically assign a unique identifier to the center of each mask and highlight each mask with a random color. These operations, in combination with our prompts, significantly enhance the generation of mask priors.
}
\label{fig:vlm_example} 
\vspace{-0.4cm}
\end{figure}
We only keep the tokens that have intersected regions with mask $\textbf{m}_{i,k}$ as the valid tokens, represented as $\tilde{{\tok}}^{proj_{i,k}}_{j}$ and $\tilde{\tok}^{rep_{j}}_{i}$. We refer readers to Fig.~\ref{fig:vggt_atth} for a comprehensive illustration.

Our main observation is that the transient query mask tends to have lower recall, which can be calculated by:
\begin{equation}
    Recall = |\tilde{\tok}^{rep_{j}}_{i}| / |\tok^{proj_{i,k}}_{j}|,
\end{equation} 
where $||$ denotes the number of tokens. By simply setting the recall threshold to 0.5, we can efficiently obtain the matching pair candidates. To further validate each pair, we use the Chamfer Distance (CD) to evaluate each pair. Using all points in the mask region is time-consuming, as the Chamfer Distance computation is relatively slow. Since the validated tokens have filtered out many irrelevant regions, we only compute the CD within the token masks. 

We adopt a score-based method for entity matching, considering only pairs with a Chamfer Distance below the threshold $Threshold_{CD}$ across all datasets. Since a lower CD indicates a more confident match, we define a normalized matching score as:
\begin{equation}
    Score = (Threshold_{CD}-CD) / Threshold_{CD}.
\end{equation}
After matching, each mask is assigned a confidence score. Masks with a score above $Threshold_{score}$ are selected as static masks, while the rest are treated as transient candidates.

\subsection{VLM Enhanced Mask Generation}
\label{VLM}
On one hand, as discussed in the introduction and confirmed by our experiments, current VLMs exhibit some zero-shot capability in identifying transient objects, but their performance is unstable—particularly when the number of transient objects is large. Therefore, directly combining VLMs with Grounding-SAM~\cite{ren2024grounded} may lead to suboptimal results. On the other hand, VGGT struggles to accurately predict geometry in large textureless regions such as sky, plain-colored curtains, or ground surfaces. Interestingly, we find that VLMs can easily interpret these regions. To leverage this complementary strength, we incorporate VLMs to enhance mask prior generation. Inspired by prior works~\cite{guo2024regiongpt}, we automatically generate labeled identifiers in large mask regions and prompt the VLM to determine whether each labeled region is static or transient. We further query the VLM to provide a concise and accurate analysis, thereby activating its Chain-of-Thought capabilities~\cite{wei2022chain, wu2025grounded}. Fig.~\ref{fig:vlm_example} illustrates the detailed VLM process.

\begin{figure*}[!ht] 
\centering
\includegraphics[width=0.98\linewidth]{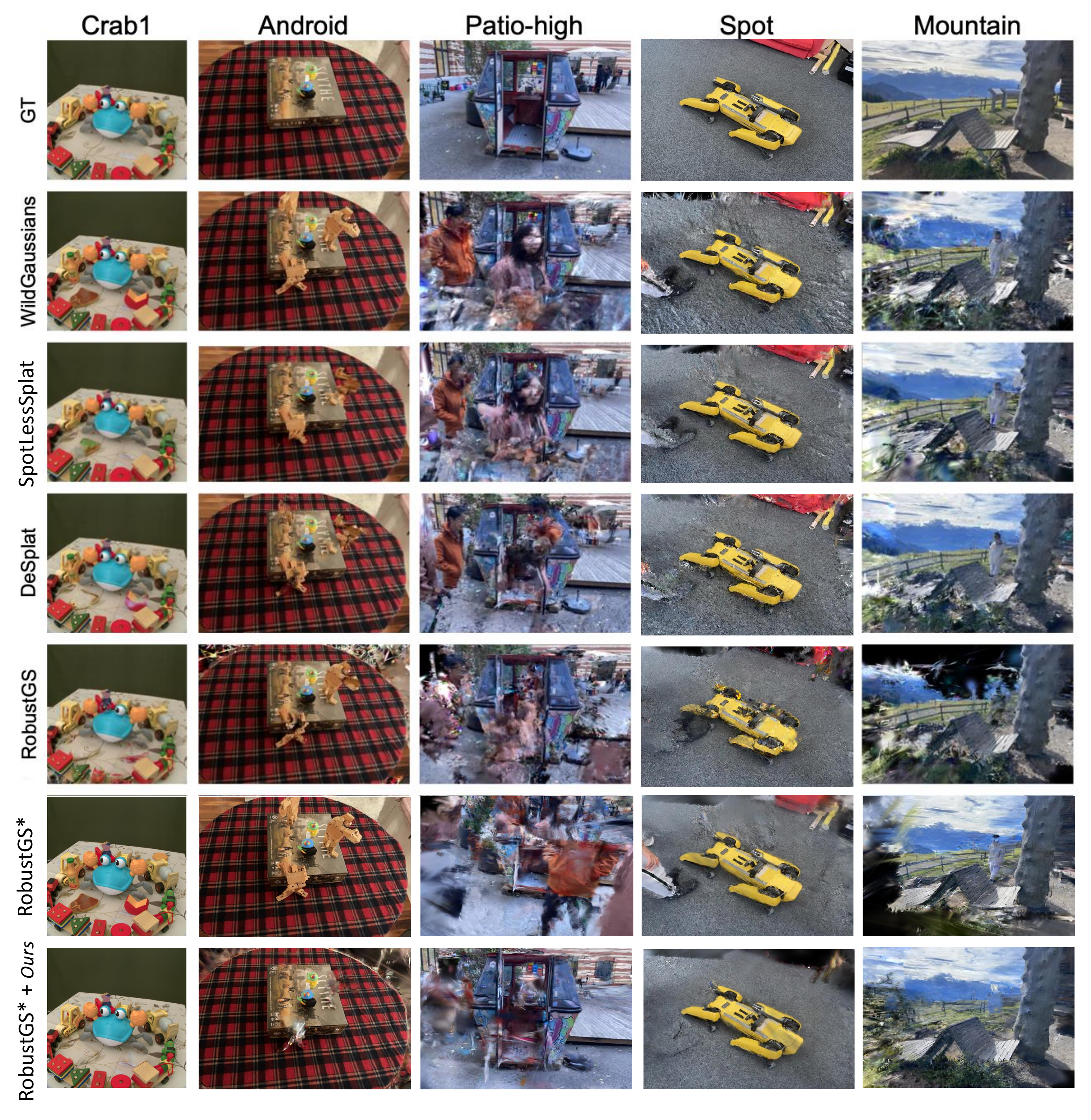}
\vspace{-0.3cm}
\caption{Qualitative evaluation of baseline methods and our approach on the NeRF On-the-go and RobustNeRF datasets. * means with the VGGT initialization.
}
\label{fig:visual_comparisons} 
\vspace{-0.4cm}
\end{figure*}

\subsection{Integrated with RobustGS}
\label{Integrate}
RobustGS~\cite{ungermann2024robust} uses logistic regression to flexibly learn the decision boundary thresholds. The input of the mask generation model is the residuals from the last iteration. Each iteration contains two back-propagation processes, one for the mask model and the other for GS parameters. RobustGS uses the masked image loss to optimize the mask generation model:
\begin{equation}
    \mathcal{L}_{mask} = \hat{\mathcal{M}} \circ \mathcal{L}_{GS} + \mathcal{L}_{reg},
\end{equation}
where $\mathcal{L}_{GS}$ contains $L$1 loss and SSIM loss and $\circ$ is the Hadamard product. $\mathcal{L}_{reg}$ is the regularization loss to avoid non-trivial solutions, such as classifying every pixel as a distractor. $\hat{\mathcal{M}}$ is the prediction of the mask model.

During the GS optimization, the mask $\hat{\mathcal{M}}$ is replaced with a non-learnable mask $\mathcal{M}$ by calculating the intersections with SAM~\cite{kirillov2023segment} predictions. Thus, the loss function in the GS optimization stage is changed into:
\begin{equation}
    \mathcal{L}^{\prime}_{mask} = \mathcal{M} \circ \mathcal{L}_{GS}.
\end{equation}
Since the mask $\hat{\mathcal{M}}$ is derived from the color residual heuristics, most regions exhibit similar residual values during the initial training stage. This makes it difficult for the simple regression model to distinguish between transient and static regions. $\mathcal{M}$, which is obtained from $\hat{\mathcal{M}}$, inherits this limitation. 

To tackle this problem, we replace $\mathcal{M}$ with our mask priors during the early stage of GS training, serving as a warm-up phase for RobustGS. We do not introduce any additional loss terms, such as an L1 loss, to explicitly align $\hat{\mathcal{M}}$ with the transient mask priors, because the priors only serve as pseudo ground truth, and the simple regression model lacks sufficient representational power.

To address the inaccuracy of the initial camera parameters, we incorporate Gaussian Bundle Adjustment (GSBA)~\cite{fan2024instantsplat} during training to further refine the camera poses. Thus, the loss of $\mathcal{L}_{GS}$ is replaced by:
\begin{equation}
    \mathcal{L}_{GSBA} = \mathcal{L}_{GS}(\{\mathcal G_i\}, \{\mathcal T_j\}),
\end{equation}
where $\mathcal T_j$ is the extrinsic of the $j$-th camera. We empirically observe that allowing camera intrinsics to be optimized or sharing camera parameters across views leads to degraded performance, often resulting in increased ghosting artifacts. Therefore, we keep the intrinsics fixed throughout training. At test time, we perform a test-time optimization where only the camera poses are updated. The initial camera parameters for the test views are obtained by rerunning VGGT on both the training and testing images.

\begin{figure}[!t] 
\centering
\includegraphics[width=0.98\linewidth]{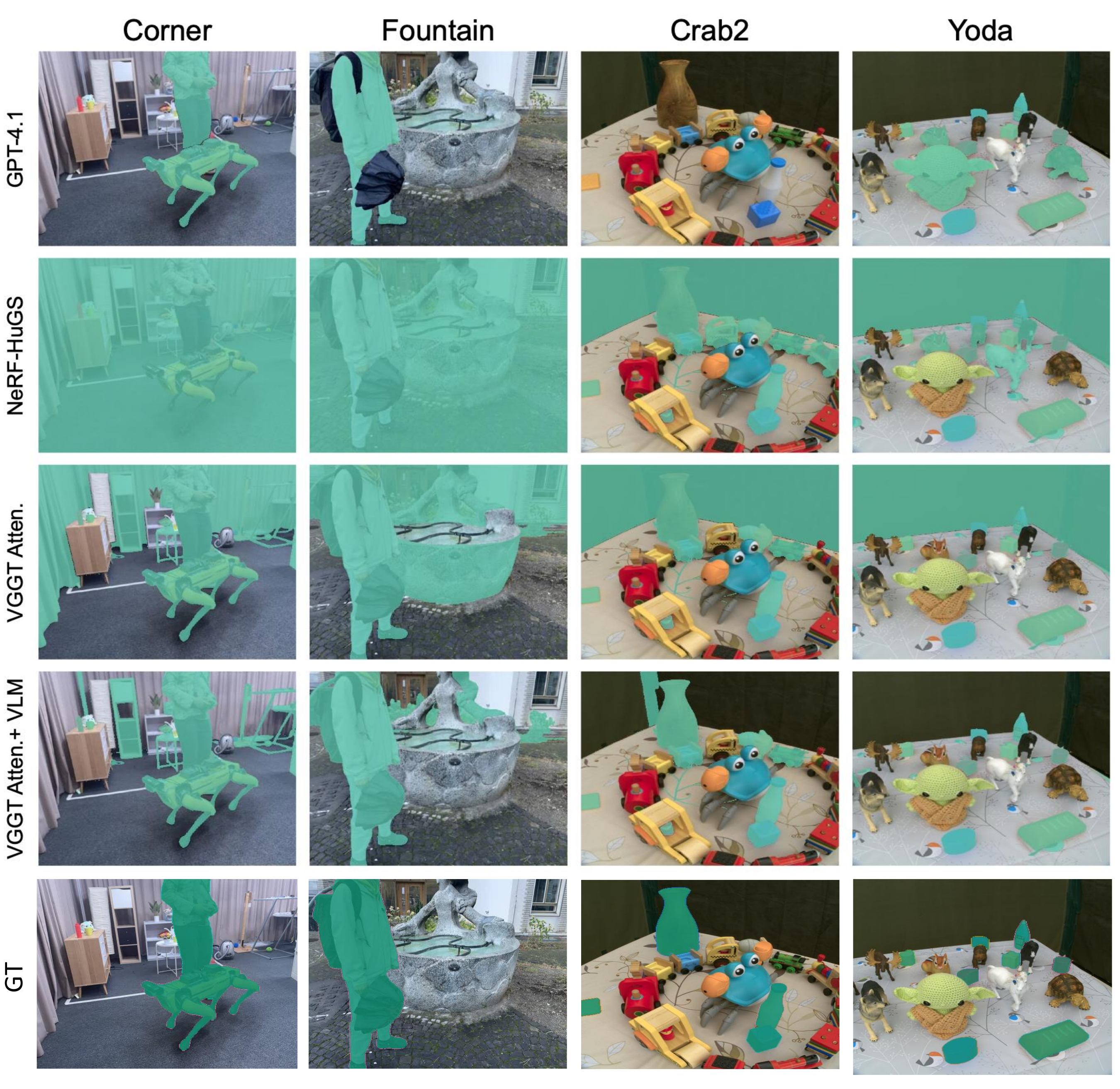}
\resizebox{1\linewidth}{!}{
\begin{tabular}{l|ccc|ccc|ccc}
\toprule
\textbf{Methods} 
& \multicolumn{3}{c|}{\textbf{Yoda}} 
& \multicolumn{3}{c|}{\textbf{Crab1}} 
& \multicolumn{3}{c}{\textbf{Crab2}} \\
& 4 views & 6 views & 8 views 
& 4 views & 6 views & 8 views 
& 4 views & 6 views & 8 views \\
\midrule
\multicolumn{10}{l}{\textbf{\textit{Closed-source VLM Models + Grounding SAM}}} \\
GPT-4.1 &  43.75 & 45.81 & 54.49 & 12.06 & 18.23 & 9.99 & 11.41 & 14.46 & 0.00 \\
GPT-4o & 32.91 & 20.15 & 50.70 & 10.35 & 2.66 & 15.33 & 1.38 & 0.00 & 0.00\\
Claude-3.7-Sonnet & 59.55 & 45.53 & 33.67 & 10.02 & 11.92 & 21.42 & 6.711 & 17.02 & 10.17\\
Moonshot-v1 & 43.40 & 21.70 & - & 6.35 & 10.01 & - & 7.69 & 15.29 & -\\
\midrule
\multicolumn{10}{l}{\textbf{\textit{Open-source VLM Models + Grounding SAM}}} \\
Qwen-VL-Max& 39.28 & 19.81 & 53.74 & 18.26 & 11.48 & 18.59 & 22.42 & 18.96 & 29.46 \\
Qwen-VL-Plus& 11.93 & 0.00 & 1.85 & 0.00 & 0.00 & 6.16 & 0.00 & 0.00 & 0.00 \\
InternVL-3 & 0.00 & 9.34 & 7.44 & 0.00 & 0.00 & 1.15 & 0.00 & 0.00 & 0.00 \\
InternVL-2.5 & 0.00 & 0.00 & 0.00 & 0.00 & 0.00 & 13.05 & 0.00 & 0.00 & 0.00 \\
\midrule
\multicolumn{10}{l}{\textbf{\textit{Geometry based methods}}} \\
NeRF-HuGS & 13.43 & 15.04 & 8.47 & 7.89 & 6.65 & 8.43 & 14.95 & 4.25 & 6.70\\
VGGT Atten. & 24.14 & 46.37 & 39.41 & 8.73 & 11.04 & 12.86 & 19.15 & 6.69 & 9.28\\
VGGT Atten. + VLM & 56.14 & 60.78 & 60.82 & 30.18 & 29.36 & 25.15 & 37.65 & 23.65 & 29.08\\
\bottomrule
\end{tabular}
}
\caption{
Quantitative and qualitative evaluation of transient mask generation. Our method consistently outperforms baseline approaches, delivering more reliable and stable mask predictions.
}
\label{fig:vis_mask_merge} 
\end{figure}

\section{Implementation, Evaluation, and Results}
\noindent
\textbf{Datasets.} We select 5 scenes from the RobustNeRF~\cite{sabour2023robustnerf} dataset and 6 scenes from the NeRF on-the-go dataset~\cite{Ren2024NeRF} to evaluate our method. Each scene consists of 6 experiments, where 4, 6, and 8 views are sampled twice. The view sampling process is implemented as a clustering procedure: the first frame is randomly selected, and subsequent views are added to the cluster only if they share more than 20 co-visible points with the already selected views.

\noindent
\textbf{Baselines and metrics.} For mask priors, we compare our method with existing VLM models coupled with Grounding-SAM~\cite{ren2024grounded} and NeRF-HuGS~\cite{chen2024nerf}. The results for NeRF-HuGS are obtained by re-running COLMAP using only the training views. We use Intersection-of-Union (IoU) to evaluate each method. Due to only the Crab and Yoda scenes containing paired photos from identical camera poses, one with distractors present and another
without and refine~\cite{ungermann2024robust} it with the 2D mask lists $\{\bm{m}_{i, k} | k = 1, ..., n_i\}$.

For distractor-free rendering, we use WildGaussians~\cite{kulhanek2024wildgaussians}, SpotLessSplat~\cite{sabour2025spotlesssplats}, RobustGS~\cite{ungermann2024robust}, and Desplat~\cite{wang2024desplat} as baseline methods. For comparison, we compute standard image reconstruction metrics (i.e., PSNR). All baseline methods are evaluated using their official implementations. Re-running COLMAP on each sparse-view scenario leads to a high failure rate, and integrating each method with GSBA is a non-trivial task. Therefore, \textbf{we use the camera parameters and initial points provided by the original densely captured datasets for all baselines.} For clarity, an asterisk (*) denotes our VGGT-based initialization in the following sections. We further add the comparison with RobustGS~\cite{ungermann2024robust} with VGGT initialization for more fair comparison.

\noindent
\textbf{Implementation details.} The $Threshold_{CD}$ is set as 0.2 across all datasets, and $Threshold_{score}$ is set as $0.5 \times N$, where $N$ is the number of training views. For VLMs enhancement, we only query the regions larger than 20,000 pixels. All prompts are defined in a general manner, without any scene-specific customization~\cite{li2025sparsegs}. We initialize the camera parameters and 3D points using VGGT, and filter the points with our high-quality mask priors. All experiments are trained for 10,000 iterations, with the first 500 iterations designated as the warm-up stage. 

\begin{table*}[!ht]   
  \caption{Quantitative results on the RobustNeRF dataset, evaluated by \textbf{PSNR}$\uparrow$, with 4, 6, 8 training views. Best results are highlighted as \colorbox[HTML]{FFCCC9}{first}, \colorbox[HTML]{FFE4CF}{second} and\colorbox[HTML]{FFFFD4}{third}.  * means with the VGGT initialization.}
  \label{tab:robustnerf}
  \centering
  \resizebox{1\linewidth}{!}{
  \begin{tabular}{@{}l|ccc|ccc|ccc|ccc|ccc}
    \toprule
     \multirow{2}{*}{\textbf{Methods}}   & \multicolumn{3}{c|}{\textbf{Android}}  & \multicolumn{3}{c|}{\textbf{Crab1}}   & \multicolumn{3}{c|}{\textbf{Crab2}}   &
     \multicolumn{3}{c|}{\textbf{Statue}}   &
     \multicolumn{3}{c}{\textbf{Yoda}}   \\
     
      &  4 views & 6 views & 8 views & 4 views & 6 views & 8 views & 4 views & 6 views & 8 views &  4 views & 6 views & 8 views &  4 views & 6 views & 8 views \\
    \midrule
    
    WildGaussians & 16.58 & 14.46 & 14.33 & 22.19 & 18.02 & 17.57 & 19.71 & 19.24 & 20.07 & 12.33 & 12.56 & 12.38 & 18.16 & 18.32 & 21.08 \\
    
    SpotLessSplat & \cellcolor[HTML]{FFCCC9}19.69 & \cellcolor[HTML]{FFCCC9}17.69 & \cellcolor[HTML]{FFCCC9}16.82 & \cellcolor[HTML]{FFCCC9}26.12 & \cellcolor[HTML]{FFE4CF}22.84 & \cellcolor[HTML]{FFE4CF}24.03 & \cellcolor[HTML]{FFCCC9}24.53 & \cellcolor[HTML]{FFCCC9}24.85 & \cellcolor[HTML]{FFCCC9}26.90 & \cellcolor[HTML]{FFFFD4}13.57 & \cellcolor[HTML]{FFFFD4}13.39 & \cellcolor[HTML]{FFE4CF}14.08 & \cellcolor[HTML]{FFCCC9}24.11 & \cellcolor[HTML]{FFCCC9}27.49 & \cellcolor[HTML]{FFCCC9}30.06 \\

    DeSplat & \cellcolor[HTML]{FFE4CF}18.71 & \cellcolor[HTML]{FFE4CF}17.24 & \cellcolor[HTML]{FFE4CF}16.10 & \cellcolor[HTML]{FFE4CF}25.00 & \cellcolor[HTML]{FFCCC9}23.13 & \cellcolor[HTML]{FFCCC9}24.83 & \cellcolor[HTML]{FFE4CF}23.69 & \cellcolor[HTML]{FFE4CF}23.31 & \cellcolor[HTML]{FFE4CF}25.75 & \cellcolor[HTML]{FFE4CF}14.15 & 13.21 & 12.60 & \cellcolor[HTML]{FFE4CF}23.36 & \cellcolor[HTML]{FFE4CF}25.04 & \cellcolor[HTML]{FFE4CF}28.62 \\

    \midrule
    
    RobustGS & 13.67 & 14.14 & 9.92 & 19.04 & 17.76 & 17.19 & 18.64 & 18.57 & 20.48 & 10.67 & 12.26 & 11.64 & 21.72 & 14.88 & 23.96 \\

    RobustGS* & 17.39 & 15.04 & 14.23 & 22.35 & 18.71 & 18.27 & 20.20 & 19.49 & 20.60 & 13.47 & \cellcolor[HTML]{FFE4CF}13.42 & \cellcolor[HTML]{FFFFD4}13.04 & 20.04 & 19.39 & 21.32 \\

    RobustGS* + \textbf{\textit{ours}}  & \cellcolor[HTML]{FFFFD4}18.00 & \cellcolor[HTML]{FFFFD4}15.85 & \cellcolor[HTML]{FFFFD4}14.88 & \cellcolor[HTML]{FFFFD4}23.81 & \cellcolor[HTML]{FFFFD4}20.23 & \cellcolor[HTML]{FFFFD4}18.99 & \cellcolor[HTML]{FFFFD4}23.17 & \cellcolor[HTML]{FFFFD4}21.51 & \cellcolor[HTML]{FFFFD4}21.58 & \cellcolor[HTML]{FFCCC9}14.38 & \cellcolor[HTML]{FFCCC9}14.25 & \cellcolor[HTML]{FFCCC9}14.37 & \cellcolor[HTML]{FFFFD4}22.36 & \cellcolor[HTML]{FFFFD4}24.99 & \cellcolor[HTML]{FFFFD4}26.55 \\ 

  \bottomrule
  \end{tabular}
  }
\end{table*}


\renewcommand{\arraystretch}{1.3}
\begin{table*}[!ht]   
  \caption{Quantitative results on the NeRF On-the-go dataset, evaluated by \textbf{PSNR}$\uparrow$, with 4, 6, 8 training views.}
  \label{tab:onthego}
  \centering
  \resizebox{1\linewidth}{!}{
  \begin{tabular}{@{}l|ccc|ccc|ccc|ccc|ccc|ccc}
    \toprule
     \multirow{2}{*}{\textbf{Methods}}   & \multicolumn{3}{c|}{\textbf{Corner}}  & \multicolumn{3}{c|}{\textbf{Fountain}}   & \multicolumn{3}{c|}{\textbf{Mountain}}   &
     \multicolumn{3}{c|}{\textbf{Spot}}   &
     \multicolumn{3}{c|}{\textbf{Patio}} & \multicolumn{3}{c}{\textbf{Patio-high}}   \\
     
      &  4 views & 6 views & 8 views & 4 views & 6 views & 8 views & 4 views & 6 views & 8 views &  4 views & 6 views & 8 views &  4 views & 6 views & 8 views &  4 views & 6 views & 8 views \\
    \midrule
    
    WildGaussians & 12.31 & 14.48 & 14.55 & 8.88 & 12.86 & 10.05 & 13.38 & 12.82 & 12.74 & 14.50 & 14.83 & 13.75 & 12.73 & \cellcolor[HTML]{FFFFD4}12.03 & 13.29 & 11.47 & 12.34 & 11.68 \\
    
    SpotLessSplat & \cellcolor[HTML]{FFFFD4}14.14 & \cellcolor[HTML]{FFCCC9}16.68 & \cellcolor[HTML]{FFCCC9}17.05 & 9.34 & \cellcolor[HTML]{FFFFD4}13.43 & 10.89 & \cellcolor[HTML]{FFE4CF}14.70 & \cellcolor[HTML]{FFE4CF}13.99 & \cellcolor[HTML]{FFE4CF}14.64 & \cellcolor[HTML]{FFFFD4}14.80 & \cellcolor[HTML]{FFE4CF}16.14 & \cellcolor[HTML]{FFFFD4}15.06 & \cellcolor[HTML]{FFE4CF}14.05 & \cellcolor[HTML]{FFE4CF}13.36 & \cellcolor[HTML]{FFFFD4}13.97 & 12.39 & \cellcolor[HTML]{FFFFD4}12.79 & \cellcolor[HTML]{FFFFD4}12.67 \\

    DeSplat & \cellcolor[HTML]{FFE4CF}14.35 & \cellcolor[HTML]{FFE4CF}15.97 & \cellcolor[HTML]{FFE4CF}16.89 & \cellcolor[HTML]{FFE4CF}10.96 & \cellcolor[HTML]{FFCCC9}13.90 & \cellcolor[HTML]{FFE4CF}11.54 & 13.58 & \cellcolor[HTML]{FFFFD4}13.86 & \cellcolor[HTML]{FFFFD4}13.71 & \cellcolor[HTML]{FFCCC9}15.25 & \cellcolor[HTML]{FFCCC9}16.64 & \cellcolor[HTML]{FFE4CF}15.42 & \cellcolor[HTML]{FFCCC9}15.14 & \cellcolor[HTML]{FFCCC9}14.13 & \cellcolor[HTML]{FFCCC9}16.10 & \cellcolor[HTML]{FFE4CF}12.97 & \cellcolor[HTML]{FFE4CF}14.05 & \cellcolor[HTML]{FFE4CF}13.58 \\

    \midrule

    RobustGS & 12.17 & 14.06 & 14.62 & 8.00 & 11.17 & 9.94 & 11.14 & 10.07 & 10.85 & 14.27 & 13.90 & 15.00 & \cellcolor[HTML]{FFFFD4}12.94 & 11.74 & \cellcolor[HTML]{FFE4CF}14.18 & 10.43 & 12.68 & 11.53 \\

    RobustGS* & 13.61 & 13.82 & 13.89 & \cellcolor[HTML]{FFFFD4}10.92 & 13.04 & \cellcolor[HTML]{FFFFD4}11.41 & \cellcolor[HTML]{FFFFD4}14.07 & 13.16 & 13.42 & 14.35 & 14.45 & 14.91 & 12.25 & 11.73 & 12.05 & \cellcolor[HTML]{FFFFD4}12.78 & 12.44 & 12.17 \\

    RobustGS* + \textbf{\textit{ours}}  & \cellcolor[HTML]{FFCCC9}14.94 & \cellcolor[HTML]{FFFFD4}14.76 & \cellcolor[HTML]{FFFFD4}15.21 & \cellcolor[HTML]{FFCCC9}11.54 &  \cellcolor[HTML]{FFE4CF}13.79 & \cellcolor[HTML]{FFCCC9}12.82 & \cellcolor[HTML]{FFCCC9}14.74 & \cellcolor[HTML]{FFCCC9}14.16 & \cellcolor[HTML]{FFCCC9}14.70 & \cellcolor[HTML]{FFE4CF}15.20 & \cellcolor[HTML]{FFFFD4}15.82 & \cellcolor[HTML]{FFCCC9}16.02 & 12.73 & 11.67 & 13.09 & \cellcolor[HTML]{FFCCC9}13.87 & \cellcolor[HTML]{FFCCC9}14.61 & \cellcolor[HTML]{FFCCC9}14.33 \\ 
    
  \bottomrule
  \end{tabular}  }
  \vspace{-0.4cm}
\end{table*}


\subsection{Mask Prior Results}
\label{mask_res}
Fig.~\ref{fig:vis_mask_merge} presents the evaluation results for prior mask generation. Overall, open-source VLMs underperform compared to their closed-source counterparts on our task. Among the closed-source models, \textit{GPT-4.1} and \textit{Claude} demonstrate the best performance, suggesting a stronger grasp of the task requirements. In contrast, the \textit{VLM + Grounding DINO} pipeline follows a cascaded approach, which is prone to error accumulation. NeRF-HuGS, which relies on COLMAP to infer static regions, often overpredicts and generally yields suboptimal results under sparse-view conditions. While VGGT attention matching alone significantly outperforms NeRF-HuGS, it still struggles in large textureless or poorly observed regions. In contrast, our method—combining VGGT with VLM guidance—consistently delivers the most stable and accurate results across all cases.



\begin{figure}[!ht]
\centering
\setlength\tabcolsep{1.8pt}
\includegraphics[width=\linewidth]{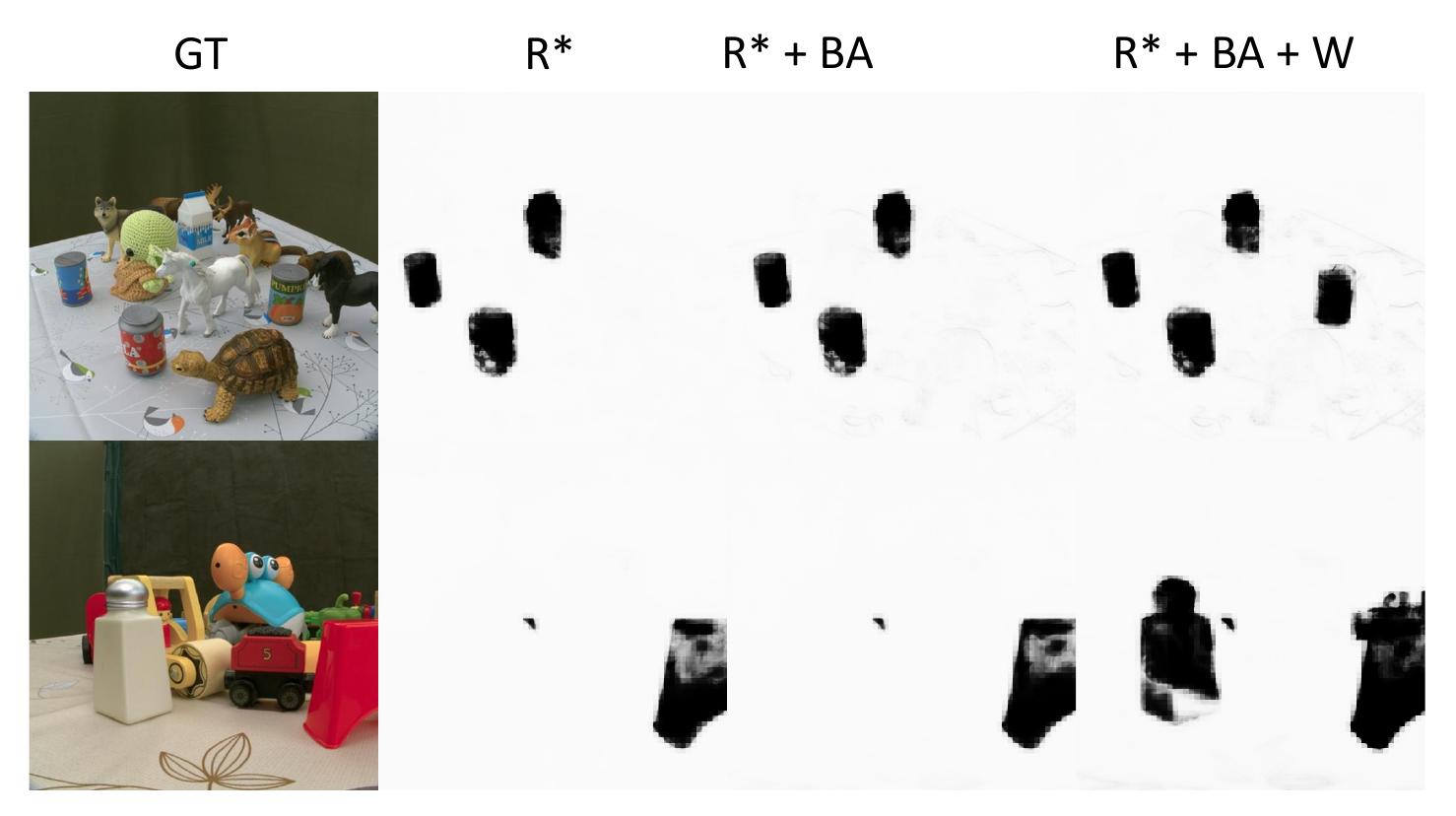}\\

\resizebox{\linewidth}{!}{
\begin{tabular}{@{}l|ccc|ccc|ccc}
    \toprule
     \multirow{2}{*}{Method}   & \multicolumn{3}{c|}{PSNR$\uparrow$}  & \multicolumn{3}{c|}{SSIM$\uparrow$}   & \multicolumn{3}{c}{LPIPS$\downarrow$} \\
      &  R* & R*+BA & Full & R* & R*+BA & Full & R* & R*+BA & Full\\
    \midrule

    Yoda   & 24.52 & 26.16 & 26.55 & 0.84 & 0.80 & 0.84 & 0.24 & 0.26 & 0.23 \\
    Crab2 & 21.22 & 21.67 & 21.58 & 0.73 & 0.76 & 0.76 & 0.32 & 0.29 & 0.29 \\
    Corner  & 14.62 & 14.98 & 15.21 & 0.43 & 0.47 & 0.48 & 0.47 & 0.44 & 0.43 \\
    Patio-high   & 13.92 & 13.75 & 14.33 & 0.32 & 0.29 & 0.34 & 0.52 & 0.53 & 0.50 \\
  \bottomrule
  \end{tabular}
} %
\captionof{figure}{
Quantitative and qualitative ablations of each component. We present the raw distractor masks generated by different ablation variants to highlight the effectiveness of our mask-prior-guided warm-up strategy.
}
\label{fig:ablation}
\vspace{-0.3cm}
\end{figure}

\begin{figure}[!ht] 
\centering
\includegraphics[width=0.98\linewidth]{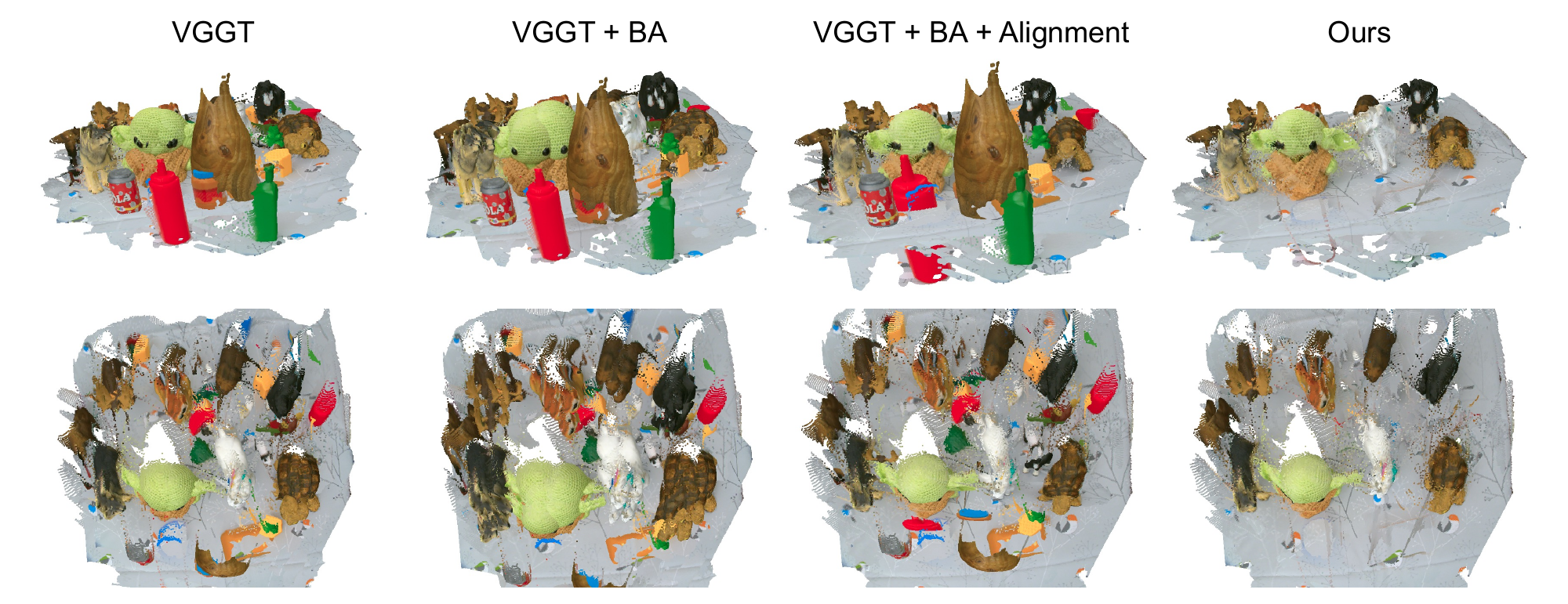}
\caption{
Illustration of VGGT combined with Bundle Adjustment (BA). The first row shows front views, while the second row presents bird’s-eye views. Our mask priors significantly enhance the robustness and reliability of BA + Alignment process. 
}
\label{fig:BA} 
\vspace{-0.4cm}
\end{figure}

\subsection{Sparse-veiw Distractor-free 3DGS Results}
\noindent
\textbf{Comparison on the RobustNeRF dataset.} As shown in Table~\ref{tab:robustnerf}, RobustGS performs the worst among all baseline methods, and WildGaussians also show relatively weak results. SpotLessSplat and Desplat are built on more advanced Gaussian Splatting frameworks with extra tailored designs and generally achieve better performance, whereas RobustGS is based on the original GS implementation. \textbf{It also should be noted that the initial parameters of baselines are drawn from the original densely captured datasets.} Despite this, our improved variant (RobustGS* + Ours) achieves competitive results, enhancing RobustGS with 2–4 dB PSNR gains and RobustGS* with 1-4 dB gains. This explicitly proves that while VGGT provides a necessary starting point, the core enhancements in rendering quality and transient removal are fundamentally driven by our high-quality mask priors and the effective warm-up strategy.


\noindent
\textbf{Comparison on the NeRF on-the-go dataset.} Similar to the trend observed on the RobustNeRF dataset, our full pipeline (RobustGS* + Ours) consistently surpasses the RobustGS and RobustGS*. As shown in Table~\ref{tab:onthego}, our method achieves the best performance in 10 out of 18 cases, outperforming all baselines.  More importantly, as demonstrated in Fig.~\ref{fig:visual_comparisons}, our robust mask priors and warm-up strategy consistently and effectively eliminate most transient objects across diverse, unconstrained scenes, proving the effectiveness and generalization ability of our method.


\noindent
\textbf{Ablation studies.} We conduct ablation studies on four scenes (Yoda, Crab2, Corner, and Spot), each with 8 input views. We compare three variants: RobustGS* (\textit{R*}), RobustGS* with GSBA (\textit{R*+BA}), and our full model (\textit{R*+BA+W}). The results are shown in Fig.~\ref{fig:ablation}. \textit{R*+BA} consistently outperforms \textit{R*}, highlighting the importance of incorporating GSBA. Our full model achieves slightly better overall performance than \textit{R*+BA} and demonstrates a clear advantage in mask generation during GS training. The qualitative parts of Fig.~\ref{fig:ablation} can better validate this advantage. The key difference lies in our mask-prior-guided warm-up strategy, which effectively reduces the burden on the mask predictor in RobustGS.

\noindent
\textbf{Discussion.} We also investigate the Bundle Adjustments before the GS training. VGGT also demonstrates that refining predicted camera poses and tracks with BA can further improve accuracy. However, the predicted depth maps are not aligned with the refined extrinsics and intrinsics.  To address this, we utilize the sparse points obtained from the BA refinement stage as pseudo-ground truth, aligning the predicted depth maps to these points using a RANSAC-based linear regression model. This alignment process is highly sensitive, particularly in scenes containing multiple transients. As shown in Fig.~\ref{fig:BA}, incorporating our mask priors can further clearly improve the success rate, demonstrating that our approach not only aids novel view synthesis but also effectively stabilizes Bundle Adjustment under challenging sparse-view and distractor-heavy conditions.



\section{Conclusion}
In this work, we present a novel method for mask prior generation to enhance distractor-free 3D Gaussian Splatting (3DGS) under sparse-view conditions. Our approach leverages the geometry foundation model (i.e., VGGT) alongside powerful Vision-Language Models (VLMs) to produce robust and reliable rendering results. With a simple mask-prior-guided warm-up strategy, we significantly boost the performance of existing distractor-free 3DGS methods.




\bibliographystyle{IEEEtran}
\bibliography{IEEEabrv,ref}

\end{document}